\documentclass[doublecol,linenumbers]{epl2} 

\usepackage{amsmath}
\usepackage{amsfonts}
\usepackage{hyperref}

\usepackage{color}
\definecolor{redcolor}{rgb}{1.0,0.,0.}

\title{Network analysis of {named entity co-occurrences} in written texts}
\shorttitle{Network analysis of named entity interactions in written texts} 

\author{Diego Raphael Amancio\inst{1}}

\institute{
  \inst{1} Institute of Mathematics and Computer Science, University of S\~ao Paulo, S\~ao Carlos, S\~ao Paulo, Brazil
}


\abstract{
{
The use of methods borrowed from statistics and physics to analyze written texts has allowed the discovery of unprecedent patterns of human behavior and cognition by establishing links between models features and language structure. While current models have been useful to unveil patterns via analysis of syntactical and semantical networks, only a few works have probed the relevance of investigating the structure arising from the relationship between relevant entities such as characters, locations and organizations. In this study, we represent entities appearing in the same context as a co-occurrence network, where links are established according to a null model based on random, shuffled texts.
Computational simulations performed in novels revealed that the proposed model displays interesting topological features, such as the small world feature, characterized by high values of clustering coefficient.
The effectiveness of our model was verified in a practical pattern recognition task in real networks. When compared with \emph{traditional word adjacency networks}, our model displayed optimized results in identifying unknown references in texts. Because the proposed representation plays a complementary role in characterizing unstructured documents via topological analysis of named entities, we believe that it could be useful to improve the characterization of written texts (and related systems), specially if combined with traditional approaches based on statistical and deeper paradigms.
}
}

\begin{document}

\maketitle

\section{Introduction}

The study of complex networks emerged in the beginning of the last decade as a powerful, robust and general representation of a myriad of real complex systems~\cite{newman2010introduction}. Biological systems, transportation networks, communication and information networks are examples of real systems whose underlying properties were unsnarled via graph representation. The investigation of many complex systems in terms of their topological structure and dynamical behavior allowed the discovery of non-trivial connectivity/functional patterns, including the discovery of specific mesoscopic, heterogeneous and hierarchical structures {responsible for particular functions~\cite{newman2010introduction}}.

Even though several complex systems modeled as networks are artlessly visualized as a graph representation (e.g. street~\cite{stretMas} or citation networks~\cite{citaphysa,Amancio2012427}), many others {undergo} a pre-processing step to allow a proper networked representation. This is the case of instances represented in an attribute space~\cite{teago}, images modelled as networks~\cite{de1000r} and texts represented as graphs~\cite{unveiling}. Of particular interest to the aims of this study are the text networks, a model that has drawn the attention of physicists in recent years and has been of paramount relevance to shed light on the structure and function of linguistic and cognitive processes~\cite{nics}. In addition, such textual representations has allowed the investigation of basic human behavior through the automatic analysis of the ever increasing amount of electronic data in social and information networks~\cite{cong2014approaching}.

Many physical models of texts have been proposed to tackle a diversity of problems~\cite{cong2014approaching}. The representation of unstructured documents depends on the specific language properties being studied. If writing style or language identification is relevant, syntactical networks are used to grasp stylistic-dependent features~\cite{i2004patterns}. In syntactical networks, each word is mapped into a node and edges are established according to {syntactical relations, which are language-dependent constraints}. Interestingly, it has been shown that such networks share the same topological properties of networks representing completely different systems~\cite{i2004patterns},
allowing thus the construction of optimized structures for language acquisition and communication~\cite{Cancho04022003}.

Usually, syntactical networks are analyzed  {using} a simplified representation, the so-called word adjacency model~\cite{prose,amancio2012complex,concentrico,teago}. In this model, adjacent words are linked if they appear as neighbors in the text. It has been shown that, despite this seemingly naive simplification, word adjacency networks {are efficient representations} for text analysis because most of the syntactical connections occur between neighboring words~\cite{i2004patterns}. The usefulness of such models {has been verified} in many theoretical and practical research~\cite{cong2014approaching}. Finally, another important class of text networks are the semantical networks, where links are established if concepts share some semantical property (e.g. a semantical similarity or a entailment relationship)~\cite{Miller:1995:WLD:219717.219748}.

{Whilst several available methods grasp the relationship between \emph{all} words or the relationship between specific classes of words~\cite{PhysRevE.74.036104,mitobolso} as a co-occurrence graph (see Section S4 of the Supplementary Information (SI)\footnote{The Supplementary Information (SI) is available from \href{https://dl.dropboxusercontent.com/u/2740286/supent.pdf}{this link}.}), only a few studies have investigated the properties of entity co-occurrences as a complex network. Note that this is a relevant issue because the presence of specific words other than relevant entities in networked models may hinder the accurate recognition of novel patterns.} In this sense, this study aims at creating a networked textual representation that analyzes the topology emerging from the relationship between specific words, {namely} the \emph{named entities},
which are a class of particular concepts that has been useful to shed light on the understanding of several language properties~\cite{statnlp}. Particularly, we consider that words represent named entities whenever they name people (characters), places {or} organizations. Hence, unlike traditional textual networked models, the proposed representation emphasizes the complexity of the document plot rather than linguistic styles.

The application of the proposed network model that links entities co-occurring in the same context allowed the identification of some interesting topological patterns. Qualitatively, we have found that {named entity networks} share topological features of other complex systems, as revealed by low values of typical shortest path length and high local clustering. In addition, we have found that the networks display a modular structure.
The potential of the proposed representation was verified in a typical pattern recognition problem devoted to the resolution of co-references. Optimized results were found with our model, which confirms that the information provided by the representation goes beyond simple word frequency statistics. Because {named entity networks} are a complementary representation of texts as networks, we believe that it could be useful to improve the characterization of written texts and related systems with nodes playing special roles.

%

\section{Extracting networks from books} \label{represent}

The objective of this paper is to propose a simple but relevant model that captures {co-occurrences of named entities} in texts. To identify pertinent entities, we used the technique referred to as ``named entity recognition'', which identifies relevant \emph{persons}, \emph{locations} and \emph{organizations} in documents. Specifically, we used the method devised in~\cite{Brill:1994}, which tags words with three possible labels. An example of entities recognition is shown in the following extract obtained from a modified Wikipedia article~\footnote{\url{en.wikipedia.org/wiki/Barack_Obama}}. Note that the entities recovered from the method are highlighted in italic font:
\vspace{-0.07em}
\ \\
\noindent
``{\emph{Barack Hussein Obama}} II (\emph{US}, born August 4, 1961) is the
44th and current President of the \emph{US}, and the first
African American to hold the office. Born in \emph{Honolulu},
\emph{Hawaii}, \emph{Obama} is a graduate of \emph{Columbia University}
and \emph{Harvard Law School}, where he served as president of the
\emph{Harvard Law Review}. He was a community organizer in \emph{Chicago}
before earning his law degree. He worked as a civil rights attorney and
taught constitutional law at \emph{University of Chicago Law School}
from 1992 to 2004.  He served three terms representing the \emph{13th
District in the Illinois Senate} from 1997 to 2004, running in 2000
unsuccessfully for the \emph{United States House of Representatives}.''
\ \\
\vspace{-0.07em}
In the above example, the recognized entities denote persons (e.g. \emph{Obama}), locations (e.g. \emph{Chicago} and \emph{Hawaii}) and organizations (e.g. \emph{Harvard Law School} or \emph{Columbia University}). After the entity recognition step, the text undergoes a pre-processing step to eliminate ambiguities and name duplications. In this step, for example, \emph{Emma Wodehouse} and \emph{Emma} are mapped into the same entity. After this pre-processing phase, a set $V = \{v_1,v_2,\ldots\}$ of entities is obtained for each book. {Note that, in this phase (training phase), only the known entities (such as \emph{Obama} and \emph{Chicago} are represented as nodes. Unknown entities (such as \emph{He}) are included in a classification phase, for example, in the anaphora resolution problem (see ``Results and discussion'' section)}.
To create the network of related entities, hereafter referred to as named entity (NE) network, the proposed method connects entities appearing in the same context. Thus, entities {sharing} some semantical relationship tend to be connected. The motivation of this approach relies on the fact that concepts appearing in the same context tend to be semantically related~\cite{modtwitter}.
The separation of the text in contexts is accomplished by splitting the entire document in shorter subtexts comprising the same number of tokens $W$. As a consequence, each book is represented by the set $\Psi = \{S_1,S_2,\ldots,S_N\}$, where $S_i$ is the $i$-th subtext.
{Note that, we could have chosen as a representative context the segments of texts structured as chapters or other natural textual structure. Specifically, we have not used this information concerning the structure of texts because this information is not readily available in all books of the dataset.}
To store the information concerning the co-occurrence of distinct entities in the same subset, the matrix $\mathbf{B}$ is created. If entity $v_i$ appears in the $j$-th subset, then $\mathbf{B}_{ij} = 1$, otherwise  $\mathbf{B}_{ij} = 0$. The frequency of entities in subsets, i.e. the number of subtexts in which a entity appears is defined as $f_i = \sum_j \mathbf{B}_{ij}$.
%
%
Analogously, the frequency of co-occurrence of two entities $v_i$ and $v_j$ is defined as
\begin{equation} \label{eq.co-occur}
    f_{ij} = \sum_k \mathbf{B}_{ik} \mathbf{B}_{jk}.
\end{equation}
The link between two entities is established if they co-occur at least in one set $S_i \in \Psi$. {The weight (i.e. the strength) of the link is computed as}
\begin{equation} \label{eqw}
    w_{ij} = \min\{P(v_i | v_j), P( v_j | v_i ) \},
\end{equation}
where $P(v_i | v_j ) = f_{ij} / f_j$. {Here}, the weight $w_{ij}$ defined in equation \ref{eqw} is used to identify the strongest links, which are  in turn stored in the adjacency matrix $\mathbf{A}$. An example of network construction for a fictitious dataset is shown in Fig. S3 of the  SI.
%

To improve the characterization of NE networks,  we also considered the significance of entities co-occurrence. More specifically, two entities $v_i$ and $v_j$ were connected only if the quantity $f_{ij}$ was sufficiently higher than the same value expected in a null model, i.e. in a random (shuffled) text.
Given two entities $v_i$ and $v_j$, the significance of the co-occurrence is estimated by computing how likely it is to observe more than $k=f_{ij}$ co-occurrences of $v_i$ and $v_j$ in the null model. Equivalently, the $p$-value associated to the quantity $k=f_{ij}$ is $p = \sum_{k \geq r} p(k)$, %
%
where $p(k)$ is the probability of $k$ co-occurrences of $v_i$ and $v_j$ in the random text.  To compute $p(k)$, we followed the approach devised in~\cite{disentangling}. If $n_1 = f_i$ and $n_2 = f_j$, $p(k)$ can be computed as
\begin{equation}  \label{feq}
p(k)= \frac{(N;k,n_1-k,n_2-k)}{(N;n_1)(N;n_2)}
\end{equation}
%
%
%
where $(x;y_1,\ldots,y_2)$  is a simplified notation:
{
\begin{equation}
    (x;y_1,\ldots,y_n) \equiv \frac{x!}{y_1! \ldots y_n!} \frac{1}{(x-y_1-\ldots y_n)!}.
\end{equation}
}
Equation \ref{feq} can be rewritten in a more convenient way, if the notation $\{a\}_b$, defined as
\begin{equation}
\{a\}_b \equiv \prod_{i=0}^{b-1} (a-i)
\end{equation}
is adopted for $a \geq b$. In this case, the likelihood $p(k)$ can be written as
\begin{eqnarray}
\centering
    p(k) &=& \frac{\{n_1\}_k \{n_2\}_k \{N-n_1\}_{n_2-k}}{\{N\}_{n_2}\{k\}_k} \nonumber \\
         &=&  \frac{\{n_1\}_k \{n_2\}_k  \{N-n_1\}_{n_2-k}}{\{N\}_{n_2-k}\{N-n_2+k\}_k\{k\}_k } \nonumber \\
    & = & \prod_{j=0}^{n_2-k-1} \Bigg{[} \frac{N - j- n_1}{N-j} \Bigg{]} \times \nonumber \\
    & & \prod_{j=0}^{k-1} \frac{(n_1-j)(n_2-j)}{(N-n_2+k-j)(k-j)}. \nonumber
\end{eqnarray}
Therefore, the $p$-value associated to the number of observed co-occurrences is~\cite{disentangling}
%
\begin{eqnarray} \label{eq.pvalue}
\centering
    p(k) &=& \sum_{k\geq r} \prod_{j=0}^{n_2-k-1} \Bigg{(} 1 - \frac{n_1}{N-j} \Bigg{)} \times \nonumber \\
    & & \prod_{j=0}^{k-1} \frac{(n_1-j)(n_2-j)}{(N-n_2+k-j)(k-j)}. \nonumber
\end{eqnarray}
%
Note that the value $p(k)$ defined in the above equation can be used to establish links between entities whose co-occurrence frequency is significant.

\section{Results and discussion} \label{osresultados}

In this section, the topological properties of NE networks are investigated. In addition, we apply the proposed networked representation to tackle a natural language processing task related to anaphora (or co-reference) resolution. The dataset used in the experiments is shown in Table S2 of the SI.
The list comprises romances by distinct authors. All books were retrieved from the Project Gutenberg dataset\footnote{\url{www.gutenberg.org}}.

\subsection{Statistical properties of named entities networks}

\begin{figure}
  \centering
  \includegraphics[width=0.45\textwidth]{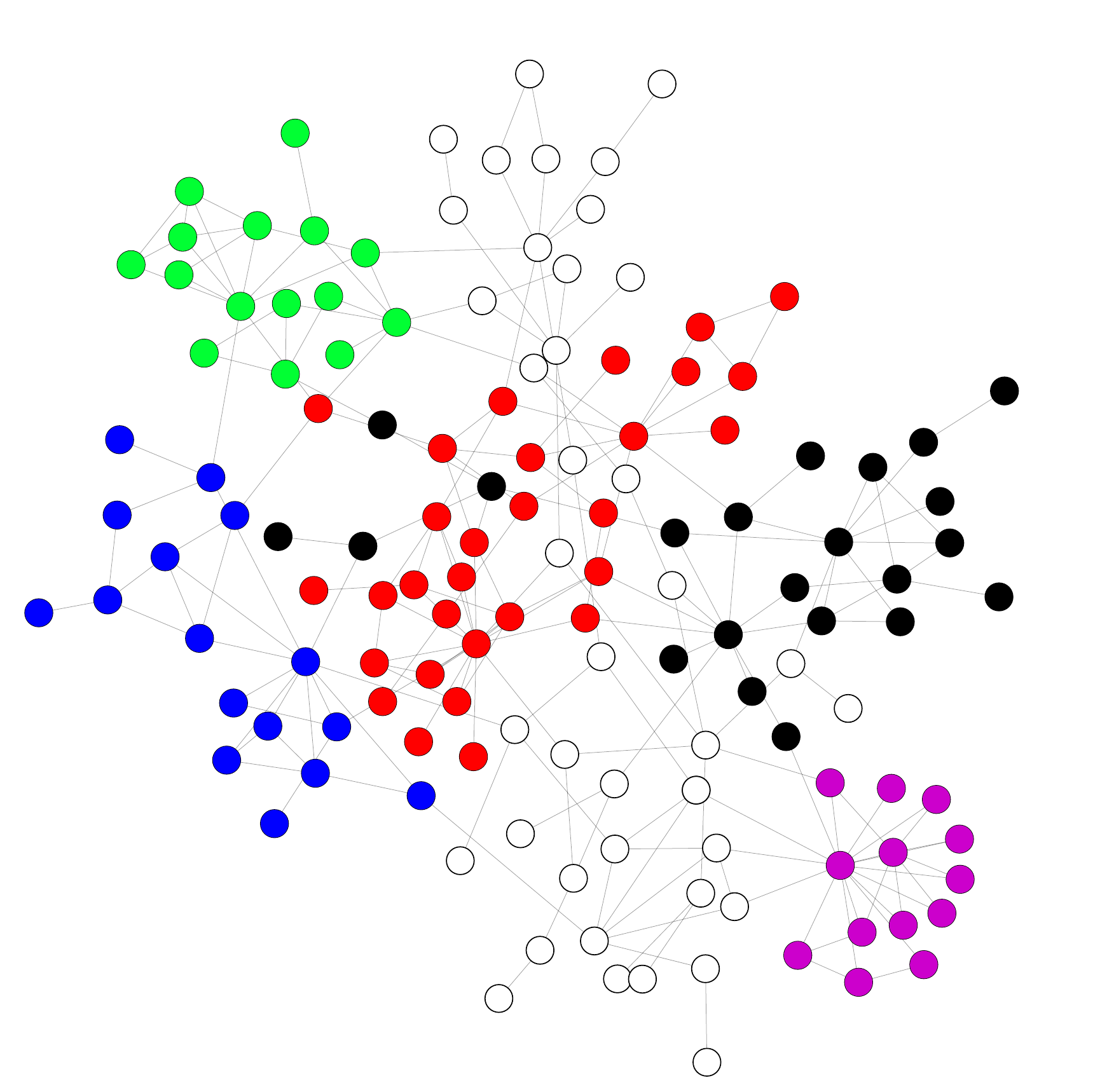}
  \caption{\label{fig.bleak}Community structure obtained from the book \emph{Bleak House}, a novel by Charles Dickens. The five largest communities are highlighted. The communities were obtained with the fastgreedy optmization of the modularity~\cite{Fortunato201075}.}
\end{figure}

To probe the topological properties of NE networks, the following quantities were computed: the number of nodes ($N$), the average degree ($\langle k \rangle$), the clustering coefficient ($\langle C \rangle$) and the average shortest path length ($\langle l \rangle$). In order to compare the properties of NE and random networks, the values of clustering coefficient and average shortest path length in equivalent random networks were also computed as:
\begin{equation}
    \langle C \rangle_r =  \langle k \rangle / N,
\end{equation}
\begin{equation}
    \langle l \rangle_r = \log N.
\end{equation}
The results obtained for selected books of our dataset are shown in Table \ref{tab.res1}. The results for the full dataset are shown in {Table S3 of the SI}.
Note that, differently from traditional language networks where the number of nodes is proportional to the vocabulary size~\cite{cnas,voynich,i2004patterns,cong2014approaching}, in this case, the total number of nodes (i.e. the number of entities) is much lower. From a qualitative point of view, NE and small-world networks share similar topological properties, because in most cases $\langle C \rangle \gg \langle C \rangle_r$ and $\langle l \rangle \simeq \langle l \rangle_r$.
{Even though the Zipf's law may play a role in the observed small-world effect, the equivalence between these two effects is not straightforward because, in the proposed model, we consider \emph{segments} of texts, where a quantity of $K$ entity occurrences does not imply the formation of $K$ edges. In addition, note that many entity occurrences tend not to be translated into an edge because of the burstiness effect, which is specially prominent in words denoting named entities~\cite{ortuno}.}
{An example of NE network is shown in Fig. S5 of the SI. As expected, central characters play a prominent role on the networked model.} Another important feature of NE networks is their modular structure. In Fig. \ref{fig.bleak}, we show the modular structure unfolded with traditional community structure methods~\cite{Fortunato201075} (see also Fig. S4 of the SI). This modular structure has been noted in most of the studied networks.
\begin{table}
\centering
\caption{\label{tab.res1}Statistical properties of NE networks. In most of the networks, the clustering coefficient is higher than the same quantity observed in equivalent random networks. With regard to the average shortest path lengths, the values observed in real and equivalent random networks are similar. {Concerning the differences across books, we have identified that most fluctuations arise from the difference in network size (result not shown). However, in specific cases (such as the value of $\langle k \rangle$ in MAN and PER), the statistical differences are significative, as a consequence of distinct {authors' styles}. Such differences could be explored in further studies aiming at identifying authorship in texts.}}
\begin{tabular}{lccccccc}
\hline
{\bf Book} & $N$ & $\langle k \rangle$ & $\langle C \rangle$ & $\langle C \rangle_{\textrm{r}}$ & $\langle l \rangle$ & $\langle l \rangle_{\textrm{r}}$ \\
\hline
MAN  & 44 & 2.68 & 0.192 & 0.062 & 2.98 & 3.83 \\
EMM       & 56 & 4.21 & 0.411 & 0.075 & 2.38 & 2.80 \\
PER & 44 & 3.27 & 0.254 & 0.074 & 3.19 & 3.19 \\
PRI      & 47 & 3.87 & 0.294 & 0.088 & 2.87 & 2.85 \\
SAS      & 33 & 3.51 & 0.204 & 0.102 & 2.80 & 2.78 \\
BLH      &140 & 3.12 & 0.241 & 0.022 & 4.90 & 4.33 \\
DCP &149 & 3.30 & 0.219 & 0.022 & 4.27 & 4.19 \\
LDR     &175 & 3.58 & 0.351 & 0.020 & 4.87 & 4.04 \\
BTW  &122 & 3.05 & 0.355 & 0.021 & 3.83 & 4.31 \\
WWL &138 & 3.65 & 0.226 & 0.026 & 4.32 & 3.81 \\
\hline
\end{tabular}
\end{table}
%
%

In order to show how the network representation might provide  complementary information for text analysis, we studied the problem of identifying the most central entities. As an example, we used the book \emph{Middlemarch}, by George Eliot. The following network centrality measurements were computed: betweenness centrality, PageRank and accessibility. The betweenness is a global measurement that computes the number of shortest paths passing through nodes. The PageRank is a global centrality measurement that considers  that a node is relevant whenever it is connected to other relevant nodes. Finally, the accessibility is a local measurement which can be understood as an extension of the degree connectivity as it quantifies the effective number of neighbors at a distance $h$ from the reference node~\cite{travenccolo2008accessibility}. Mathematically, the accessibility ($\alpha$) is defined as
\begin{equation}
	\alpha^{(h)}(i) = \exp \Big{(} - \sum p^{(h)}_{ij} \ln p^{(h)}_{ij} \Big{)} ,
\end{equation}
where $p^{(h)}_{ij}$ is the probability of a random walker to go from node $i$ to node $j$ in $h$ steps. All three network centrality indices were compared with the raw frequency of appearance of the respective entities in the books. {In Table \ref{tab.centr}, we show in decreasing order the 10 most relevant entities obtained with the frequency and the accessibility measurement. The most relevant entities obtained with the betweenness and PageRank are shown in Table S5 of the SI}.
As for the betweenness centrality, the six most relevant entities are the same as the ones obtained with the simple frequency. However, the entities appearing between seventh and tenth positions are low frequency entities that, nonetheless,  are located in privileged network position. A similar behavior can be observed for the PageRank index: top relevant entities are also very frequent, even though some low frequency entities are important because they are a few hops away from the most relevant nodes. The {most relevant entities} captured by the accessibility index (computed for $h=2$) turned out to be less influenced by the frequency.  Note, for example, that \emph{Farebrother} is the character with the highest effective number of  neighbors at the second level, even though it is only the tenth most frequent entity in the book. {In fact, a correlation analysis of less frequent entities showed that the relevance of entities according to topological indices may not be predicted by frequency alone (result not shown). By no means we are suggesting that network measurements are more relevant than traditional frequency indices. We rather suggest that network measurements could be included as an additional feature to analyze e.g. the complexity of book plots and literary movements~\cite{literary,avancos,xu,masucci}.}

%

\begin{table}
\centering
\caption{\label{tab.centr}{Rank of most relevant entities in Middlemarch (a novel by George Eliot) according to two centrality measurements}. While frequency is obtained from  simple word counts, the accessibility measurement is obtained from NE networks. Note that the most frequent entities assume different levels of relevance in the network according to the accessibility. }
\begin{tabular}{|c|l|l|}
\hline
\#& {\bf Frequency} & {\bf Accessibility}  \\
\hline
1& Tertius Lydgate &  Camden Farebrother \\
2& Dorothea Brooke &  Tertius Lydgate \\
3& Edward Casaubon &  Rosamond Vincy\\
4& Fred Vincy      &  Mr. Tyke \\
5& Rosamond Vincy  &  Edward Casaubon \\
6& Nicholas Bulstrode  & Dorothea Brooke\\
7& Mary Garth      &  Caleb Garth \\
8& Celia Brooke    &  Nicholas Bulstrode \\
9& James Chettam   & Fred Vincy \\
10& Candem Farebrother & Laure \\
\hline
\end{tabular}
\end{table}


\subsection{Identifying patterns in NE networks: application to anaphora resolution} \label{anafora}

To show how the proposed network representation might be useful in a real application, we addressed the anaphora (or co-reference) resolution problem~\cite{Zheng20111113} ({see a brief description of traditional methods in Section S4 of the SI}). In this task, we aim at identifying the entity related to {an undefined} reference in the text. To illustrate the problem, consider the following example:
``\emph{The ordinance was published by Ricardo and Borsari, who are responsible for the management of water resources in the state. In the document, he points out the critical situation of water in the region.}''
{Note that, in this case, the objective of a system aiming at anaphora resolution is to relate ``\emph{who}'' and ``\emph{he}'' to either ``\emph{Ricardo}'' or ``\emph{Borsari}''.}

The following methodology was applied to tackle the anaphora resolution problem. {Differently from the training phase (see Section ``Extracting networks from books''), in this phase (classification phase), each unknown entity (e.g. ``\emph{who}'' or ``\emph{he}'' in the previous example) was modelled as a different node in the network. As such, the network comprises nodes belonging to the set of unknown ($E_?$) and known ($E_!$) entities. In other words, the final networks in this task contain vertices of different linguistic provenance, {i.e. while some vertices represent known named entities; others model unknown references.}
We measured all possible pairwise similarities between entities $v_i \in E_?$ and $v_j \in E_!$. Therefore, unknown references {were characterized} by their similarity to the others entities in the book. In the baseline approach based on simple co-occurrence statistics (COO), which does not consider the global topology of networks, each $v_i \in E_?$ is considered to be similar only to those entities $v_j \in E_!$ that appeared in the same text window. Mathematically, the similarity $f_{ij}$ between $v_i \in E_?$ and $v_j \in E_s$ in the baseline approach is computed as defined in equation \ref{eq.co-occur}.

{
The co-occurrence approach is based on the premise that the references to unknown entities tend to occur in the same context, thus surrounding characters, locations and organizations tend to be the same. However, the co-occurrence approach only takes into account local information to measure the similarity between entities, which may lead to a large loss of relevant information. A more informed approach can consider the full network topology to quantify the pairwise similarities and thus improve the characterization of the context around a entity.}
Here, we used the Katz similarity~\cite{newman2010introduction}, which is defined as:
\begin{equation} \label{eq.katz}
	\mathbf{\kappa} = \sum_{i=0}^\infty \alpha^m \mathbf{A}^m = (\mathbf{I}-\alpha \mathbf{A})^ {-1},
\end{equation}
where $\mathbf{I}$ is the identity matrix and $\alpha$ is a positive constant. If $\lambda_1$ is the leading eigenvalue of $\mathbf{A}$, $\alpha$ must satisfy $\alpha < \lambda_1^{-1}$ if equation \ref{eq.katz} is to converge~\cite{newman2010introduction}. We also considered a variation of the Katz similarity ($\tilde{\kappa}$) that does not consider the bias toward highly connected nodes. Mathematically, the similarity $\tilde{\kappa}$ between two nodes $v_i \in E_?$ and $v_j \in E_!$ is given by:
\begin{equation}
  \tilde{\kappa}_{ij} = \frac{\alpha}{f_i} \sum_k \mathbf{A}_{ik} \tilde{\kappa}_{kj} +  \delta_{ij},
\end{equation}
where $\delta_{ij}$ accounts for the self-similarity term and is defined as $\delta_{ij} = 1$, if $v_i = v_j$; and $\delta_{ij} = 0$, otherwise.
%
In matrix terms, $\tilde{\kappa}$ is written as
\begin{equation} \label{eq.nkatz}
	\tilde{\kappa} = (\mathbf{D} - \alpha \mathbf{A})^{-1} \mathbf{D},
\end{equation}
where $\mathbf{D}$ is the diagonal matrix with elements $\mathbf{D}_{ij} = f_i$, if $v_i = v_j$; and $\mathbf{D}_{ij} = 0$; otherwise.
%
Note that both measurements $\kappa$ and $\tilde{\kappa}$ make use of the full network structure to compute pairwise similarities. This is evident when one interprets the quantity $\mathbf{A}^m$ as the matrix storing the number of paths of length $m$ between two nodes~\cite{newman2010introduction}. An example of identification of unknown entities in a {toy network} is discussed in {Section S3 of the SI.}

In our real dataset, we addressed the anaphora resolution task where two possible entities are candidates in a unsolved reference. This problem can be modelled as a supervised classification task~\cite{systematic} with two possible classes ({see definitions in Section S1 of the SI}). The pairs of evaluated entities  are shown in the second column of Table \ref{tab.anaph}. The best average accuracy rates obtained in our selected dataset are shown in Table \ref{tab.anaph}. For each pair of entities, we show the performance obtained with the traditional method based on co-occurrence statistics (COO, see equation \ref{eq.co-occur}) and with methods making use of network similarity measures, as defined in equation \ref{eq.katz} (Katz similarity). The results obtained with the normalized Katz similarity (see equation \ref{eq.nkatz}) were no better than the ones obtained with the non-normalized version, {as shown in Table S4 of the SI}. Note that the {use of global network information} was able to improve the characterization of unsolved references, as revealed by higher accuracy indexes obtained with {non-local} network measurements when compared to the traditional method based on simple co-occurrence statistics. {This means that the co-occurrence with specific entities might be useful to discriminate unknown entities whenever the test instance tend to appear in a community dominated by a specific entity. In fact, a systematic error analysis performed in our dataset revealed that most of the errors occur when \emph{distinct} entities in the training dataset appear in the \emph{same} community. Another recurrent error occurs when the test instance appears between two communities dominated by distinct entities (see Section S5 of the SI).}
This result confirms the importance of the patterns unveiled by the proposed networked representation, which were hidden from traditional models.


\begin{table}
\centering
\caption{\label{tab.anaph}Accuracy rate obtained when identifying references using the traditional model based on simple co-occurrence statistics (COO, as defined in equation \ref{eq.co-occur}) and the proposed networked approach (see equation \ref{eq.katz}). {The results obtained with equation \ref{eq.nkatz} are shown in Table S4 of the SI. The results obtained with traditional co-occurrence techniques not relying on networked information are shown in Table S6 of the SI.}}
\begin{tabular}{|l|l|c|c|}
\hline
{\bf Book} & {\bf Entities} &  {\bf COO}  & {\bf Katz}  \\
\hline
BLH & Esther and Caroline    & 69.0\%  &  81.7\%  \\
BLH & Richard and Ada        & 58.0\%  &  75.0\%  \\	
EMM & Emma and Harriet       & 51.7\%  &  86.7\%  \\
EMM & Emma and Jane          & 66.0\%  &  86.0\%  \\
JUD & Jude and Richard       & 59.2\%  &  87.7\%  \\
LDR & Arthur and Mr. Pancks  & 55.9\%  &  81.7\%  \\	
MMA & Edward and Dorothea 	 & 50.0\%  &  87.0\%  \\
MMA & Rosamond and Lydgate   & 52.0\%  &  84.8\%  \\
WWL & Felix and Lady Carbury & 64.0\%  &  84.0\%  \\	
WWL & Francis and Eleanor    & 52.0\%  &  80.0\%  \\	
\hline
\end{tabular}
\end{table}

{A systematic comparison with other traditional anaphora resolution methods not relying on any network information was also performed, and the results are provided in Table S6 of the SI. In general, our networked approach performed better than other statistical approaches. In light of the results obtained with NE networks drawing on the Katz measurement to quantify similarity between unknown references, we advocate that topological information could be used to complement more informed approaches based on deep language analysis.}
\vspace{-0.25em}
\section{Conclusion} \label{aconclusao}

In this study, we have introduced a model to form named entities networks, i.e. networks whose nodes denote people, locations and organizations. Unlike current networked representations of written texts, our model focused on the complexity arising from non-trivial {co-occurrences} between entities alone. In other words, we disregarded linguistic/stylist influences of the language on the construction of the networks. From this point of view, NE networks can be understood as a complementary form of text representation. A topological analysis performed on novels revealed that NE networks display high-clustering and low typical shortest path lengths, a similar behavior found in other textual and non-textual networks. Another interesting finding arising from the topological analysis of our model is that NE networks can unveil patterns that cannot be unveiled with traditional methods based on simple word count statistics. This was clear in the application of NE networks for identifying unknown references in texts. Particularly, the characterization textual relying upon NE networks outperformed the traditional representation based on statistical co-occurrence analysis.
{We believe that the performance of applications using NE networks could be improved with further development of enhanced automatic entity recognizers, which could treat referential opacities~\cite{opacidade} and entities with multiple aliases (e.g. ``morning star'' and ``evening star'')~\cite{mitkov}}.

NE networks turned out to play a complementary role in the characterization of written texts. While traditional approaches neglect the rich information underlying networked representations, our model is able to capture this type of information concerning the interactions between relevant entities. Given the variety of current applications and representations making use of networks in texts~\cite{cong2014approaching}, we believe that the use of NE networks might be useful to recover language-independent patterns. From a practical point of view, the model could also lead to improved performances for recognizing styles, authorship, quality and plagiarisms in a higher level of abstraction.

\acknowledgments
This work was supported by FAPESP (grant no. 14/20830-0).
I thank Vanessa Queiroz Marinho and Filipi Nascimento Silva for a careful reading of the manuscript.

\end{document}